\documentclass{INTERSPEECH2023}

\interspeechcameraready

\title{Open Universal Arabic ASR Leaderboard}
\name{Yingzhi Wang$^1$, Anas Alhmoud$^1$, Muhammad Alqurishi$^1$}
\address{$^1$ELM Company, KSA}
\email{wangyingzhi666@gmail.com, aalhmoud@elm.sa, mualqurishi@elm.sa}

\begin{document}

\maketitle

\begin{abstract}
In recent years, the enhanced capabilities of ASR models and the emergence of multi-dialect datasets have increasingly pushed Arabic ASR model development toward an all-dialect-in-one direction. This trend highlights the need for benchmarking studies that evaluate model performance on multiple dialects, providing the community with insights into models' generalization capabilities.

In this paper, we introduce Open Universal Arabic ASR Leaderboard, a continuous benchmark project for open-source general Arabic ASR models across various multi-dialect datasets.
We also provide a comprehensive analysis of the model's robustness, speaker adaptation, inference efficiency, and memory consumption.
This work aims to offer the Arabic ASR community a reference for models' general performance and also establish a common evaluation framework for multi-dialectal Arabic ASR models.



\end{abstract}

\section{Introduction}

Arabic Automatic Speech Recognition (ASR) research faces unique challenges compared to other languages like English. The complexity of Arabic, with its rich morphology, dialectal variations, and lack of diacritics in written text, poses difficulties for both data collection and model development \cite{review1, review2}. Particularly regarding dialects, while Modern Standard Arabic (MSA) is commonly used in formal settings, there exist numerous regional dialects with notable differences in both pronunciation and vocabulary.
Despite the challenge of developing an ASR model that generalizes well across all Arabic dialects, there has been a growing emergence of multi-dialect Arabic ASR models in recent years. This is driven by the collection of extensive multi-dialect datasets, enhancements in ASR model capabilities, and the growing prevalence of speech toolkits and communities.

Multi-dialect datasets play a crucial role in developing universal Arabic ASR models.
Arabic dialects include but are not limited to Egyptian, Levantine, Gulf, and Maghrebi, each with distinct phonological, lexical, and syntactic variations. In this situation, gathering and labeling data require considerable time and effort. However,
over the past few years, multi-dialectal datasets such as \cite{mgb2, sada, masc, qasr, casablanca} have become increasingly available.
For example, MGB-2 \cite{mgb2} consists of transcribed audio recordings sourced from Arabic broadcast TV programs, spanning over 1,200 hours of content and covering MSA, Egyptian, Gulf, Levantine, and North African dialects. The QASR dataset \cite{qasr} includes over 2000 hours of speech samples collected from diverse sources, including broadcast media and conversational contexts, and provides labeled data for both MSA and prominent Arabic dialects. These datasets offer rich linguistic resources that capture the nuances of regional dialects, laying the groundwork for multi-dialectal ASR model training.

Besides, End-to-end ASR models have made remarkable progress in recent years, driven by innovations in both model architectures and training techniques. Self-supervised models \cite{ssl-review}, such as Wav2Vec 2.0 \cite{wav2vec2} and HuBERT \cite{hubert}, enable representation learning on vast amounts of unlabeled data, followed by fine-tuning on limited labeled datasets to enhance performance in various downstream tasks. Conformer encoders \cite{conformer}, combining the strengths of convolutional and transformer architectures, have also been widely adopted for their ability to model both local and global dependencies in speech sequences. Additionally, OpenAI's Whisper series \cite{whisper} has pushed ASR capabilities further by employing large-scale transformer-based models together with large-scale labeled datasets. These innovations support the development of multi-dialectal models from the perspective of modeling capabilities.

Speech AI toolkits and communities are also important in advancing speech processing technologies by providing powerful frameworks and platforms for research and development. For instance, HuggingFace \cite{huggingface} offers a wide range of pre-trained models and easy-to-use tools for speech tasks, fostering a vibrant community of developers and researchers. NVIDIA NeMo \cite{nemo}, with its focus on scalable, GPU-accelerated models, provides robust ASR frameworks and pre-trained models. SpeechBrain \cite{speechbrain}, the all-in-one toolkit designed for a wide range of speech processing tasks, also provides a flexible and modular framework for developing and experimenting with state-of-the-art models. These toolkits and platforms make state-of-the-art models accessible and easy to use for researchers and developers, which has greatly accelerated the development of Arabic ASR models.



With the growth of multi-dialect ASR models, there is an increasing demand for a thorough benchmark to assess their performance. However, existing benchmark studies are limited in scope, as they encompass limited model types and seldom employ multi-dataset strategies, which restricts the generalizability of their results. For example, in \cite{casablanca}, 4 models were evaluated on a single multi-dialect dataset. \cite{n-shot} evaluates whisper and wav2vec XLS-R \cite{xlsr} models in n-shot setting on Common Voice\cite{commonvoice}, MGB-2\cite{mgb2}, MGB-3\cite{mgb-3}, MGB-5\cite{mgb-5} and Fleurs\cite{fleurs} datasets. However, MGB-3 and Fleurs contain only Egyptian dialects while MGB-5 contains only the Moroccan dialect, this limited coverage restricts the evaluation of the models' generalizability against the diversity of Arabic dialects. \cite{larabench} evaluates 4 models on 7 Arabic datasets, 5 of which are mono-dialectal. In contrast to Arabic, the English ASR community benefits from a comprehensive leaderboard \cite{leaderboard} featuring a wide range of models and multiple main-stream datasets, which has garnered substantial attention and serves as a key reference in the ASR community.

In this work, inspired by \cite{leaderboard}, we address the existing gap by introducing the largest Arabic ASR benchmark to date. Our benchmark leverages a multi-dataset strategy with exclusively multi-dialectal datasets, effectively demonstrating the model's generalizability across diverse dialects. It covers most of the state-of-the-art Arabic ASR models, offering insights for different research and production environments.

\section{Experimental Settings}
We conducted zero-shot inference for a diverse set of popular open-source Arabic ASR models on a collection of multi-dialectal Arabic datasets. 

\subsection{Models}
To facilitate the collection of state-of-the-art Arabic ASR models, we divided them into the following four main categories.

\textbf{Whisper}: The OpenAI's Whisper series of models have demonstrated strong capabilities across multiple languages and diverse recording environments. The training set scales to 680,000 hours of multilingual and multitask supervision of which 739 hours are labeled Arabic speech. The model architecture is the transformer-based encoder-decoder and is scaled to 1550M parameters for large models. In our experiments we involved whisper-large\footnote{\url{https://huggingface.co/openai/whisper-large}}, whisper-large-v2\footnote{\url{https://huggingface.co/openai/whisper-large-v2}}, whisper-large-v3\footnote{\url{https://huggingface.co/openai/whisper-large-v3}} and whisper-large-v3-turbo\footnote{\url{https://huggingface.co/openai/whisper-large-v3-turbo}}. Whisper-large is the original largest version of the model, while the whisper-large-v2 model is trained for 2.5x more epochs with added regularization for improved performance. Whisper-large-v3 improves upon v2 by using 128 Mel frequency bins (up from 80) and training on a larger dataset, leading to a 10-20\% reduction in errors across various languages. Whisper-large-v3-turbo is a pruned, fine-tuned version of Whisper-large-v3, reducing decoding layers from 32 to 4, significantly boosting speed with only minor quality degradation. For comparison in low-resource experimental settings, we also included Whisper-medium\footnote{\url{https://huggingface.co/openai/whisper-medium}} and Whisper-small\footnote{\url{https://huggingface.co/openai/whisper-small}}.

\textbf{Conformer}: Conformer is another highly efficient deep learning model architecture designed for speech recognition tasks. It combines convolutional neural networks (CNNs) and transformers to capture both local and global features of speech. NVIDIA NeMo provides performant conformer-based Connectionist Temporal Classification (CTC) \cite{ctc} and RNN-Transducer (RNNT) \cite{rnnt} models pre-trained on different languages. In our experiment, we included Nvidia's Conformer-CTC-large-Arabic\footnote{\url{https://catalog.ngc.nvidia.com/orgs/nvidia/teams/riva/models/speechtotext_ar_ar_conformer}} which has around 120M parameters and is pre-trained on ASRSet 3.0 \cite{asrset} with over 3600 hours of Arabic speech. Together with the Conformer-CTC model, NeMo also releases a 4-gram Arabic language model \footnote{\url{https://catalog.ngc.nvidia.com/orgs/nvidia/teams/riva/models/speechtotext\_ar\_ar\_lm}}. Therefore, we tested both greedy decoding and beam search with this language model.

\textbf{Self-supervised models}:
Self-supervised models like wav2vec 2.0 and HuBERT have demonstrated state-of-the-art performance in ASR by fine-tuning on sufficient labeled data. As for Arabic ASR, due to the lack of officially released fine-tuned models, we also gathered popular and performant models from the community and included them in our benchmark. We tried to cover different types of self-supervised models to maximize model diversity. The model names mentioned below are taken from the Hugging Face model cards. \textit{speechbrain/asr-wav2vec2-commonvoice-14-ar}\footnote{\url{https://huggingface.co/speechbrain/asr-wav2vec2-commonvoice-14-ar}} is a fine-tuned wav2vec 2.0 on Common Voice 14.0 released by SpeechBrain. \textit{jonatasgrosman/wav2vec2-large-xlsr-53-arabic}\footnote{\url{https://huggingface.co/jonatasgrosman/wav2vec2-large-xlsr-53-arabic}} is a fine-tuned version of wav2vec 2.0 XLS-R \cite{xlsr} for Arabic ASR using Common Voice 6.1 and Arabic Speech Corpus \cite{asc}. \textit{facebook/mms-1b-all}\footnote{\url{https://huggingface.co/facebook/mms-1b-all}} is an officially released pre-trained wav2vec 2.0 model supporting multilingual ASR for over 1000 languages\cite{mms}, for the evaluation we used its Arabic ASR adaptor. \textit{whitefox123/w2v-bert-2.0-arabic-4}\footnote{\url{https://huggingface.co/whitefox123/w2v-bert-2.0-arabic-4}} is a fine-tuned wav2vec-bert 2.0 \cite{w2v-bert} and \textit{asafaya/hubert-large-arabic-transcribe}\footnote{\url{https://huggingface.co/asafaya/hubert-large-arabic-transcribe}} is a fine-tuned HuBERT.

\begin{table*}[htbp]
  \caption{The Open Universal Arabic ASR Leaderboard. Apart from the WER/CER for each test set, we also report the Average WER/CER and rank the models based on the Average WER, from lowest to highest.}
  \label{tab:results}
  \centering
  \resizebox{\textwidth}{!}{
  \begin{tabular}{c c c c c c c c c c c c c c}
    \toprule

    & & \multicolumn{2}{c}{Average} & \multicolumn{2}{c}{SADA(Test)} & \multicolumn{2}{c}{CV-18.0(Test)} & \multicolumn{2}{c}{MASC(Test-Clean)} & 
    \multicolumn{2}{c}{MASC(Test-Noisy)} & \multicolumn{2}{c}{MGB-2(Test)} \\
    model & rank & WER$\downarrow$ & CER & WER & CER & WER & CER & WER & CER & WER & CER & WER & CER \\
    \midrule
    nvidia/conformer-ctc-large-ar (lm) & 1 & \textbf{25,71\%} & 10,02\% & \textbf{44,52\%} & 23,76\% & 8,80\% & 2,77\% & \textbf{23,74\%} & \textbf{5,63\%} & 34,29\% & 11,07\% & 17,20\% & \textbf{6,87\%} \\

    nvidia/conformer-ctc-large-ar (greedy) & 2 & 27,46\% & \textbf{9,94\%} & 47,26\% & \textbf{22,54\%} & 10,60\% & 3,05\% & 24,12\% & 5,63\% & 35,64\% & \textbf{11,02\%} & 19,69\% & 7,46\% \\

    openai/whisper-large-v3 & 3	& 29,87\% & 13,65\% & 55,96\% & 34,62\%	& 17,83\% & 5,74\% & 24,66\% & 7,24\% & 34,63\% & 12,89\% & \textbf{16,26\%} & 7,74\% \\

    facebook/seamless-m4t-v2-large & 4 & 32,55\% & 14,47\% & 62,52\% & 37,61\% & 21,70\% & 6,24\% & 25,04\% & 7,19\% & \textbf{33,24\%} & 11,92\% & 20,23\% & 9,37\% \\

    openai/whisper-large-v3-turbo & 5 & 33,30\% & 15,68\% & 60,36\% & 37,67\% & 25,73\% & 10,89\% & 25,51\% & 7,55\% & 37,16\% & 13,93\% & 17,75\% & 8,34\% \\

    openai/whisper-large-v2 & 6 & 34,04\% & 16,26\% & 57,46\% & 36,59\% & 21,77\% & 7,44\% & 27,25\% & 8,28\% & 38,55\% & 15,49\% & 25,17\% & 13,48\% \\

    openai/whisper-large & 7 & 36,65\% & 17,44\% & 63,24\% & 40,16\% & 26,04\% & 9,61\% & 28,89\% & 9,05\% & 40,79\% & 16,31\% & 24,28\% & 12,10\% \\

    asafaya/hubert-large-ft & 8	& 39,29\% & 13,61\% & 67,82\% & 31,83\% & \textbf{8,01\%} & \textbf{2,37\%} & 32,94\% & 7,15\% & 50,16\% & 15,62\% & 37,51\% & 11,07\% \\

    openai/whisper-medium & 9 & 39,60\%	& 19,10\% & 67,71\% & 43,83\% & 28,07\% & 10,38\% & 29,99\% & 8,98\% & 42,91\% & 17,49\% & 29,32\% & 14,82\% \\

    facebook/mms-1b-all & 10 & 47,86\% & 17,66\% & 77,48\% & 37,50\% & 26,52\% & 7,21\% & 38,82\% & 10,36\% & 57,33\% & 19,76\% & 39,16\% & 13,48\% \\

    whitefox123/w2v-bert-2.0-ft & 11 & 48,62\% & 16,79\% & 78,02\% & 33,17\% & 24,18\% & 6,79\% & 35,93\% & 9,01\% & 56,36\% & 19,43\% & 48,64\% & 15,56\% \\ 

    openai/whisper-small & 12 & 52,18\% & 25,15\% & 87,34\% & 56,75\% & 41,79\% & 15,75\% & 37,82\% & 11,92\% & 53,28\% & 21,93\% & 40,66\% & 19,39\% \\
    
    jonatasgrosman/wav2vec2-large-xlsr-ft & 13 & 54,63\% & 21,46\% & 86,82\% & 44,20\% & 23,00\% & 6,64\% & 42,75\% & 11,87\% & 64,27\% & 24,17\% & 56,29\% & 20,44\% \\

    speechbrain/wav2vec2-large-ft & 14 & 60,15\% & 26,64\% & 88,54\% & 50,28\% & 29,17\% & 9,85\% & 49,10\% & 16,37\% & 69,57\% & 30,17\% & 64,37\% & 26,56\% \\
    
    \bottomrule
    \end{tabular}
}
\end{table*}

\textbf{Multi-Task Foundation Models}: 
In addition to the models mentioned above,
we also considered multi-task and even multi-modal foundation models. However, currently, this kind of model offers little support for Arabic ASR. In our experiment, we included Meta's Seamless M4T \cite{seamless}\footnote{\url{https://huggingface.co/facebook/seamless-m4t-v2-large}}, which supports Arabic ASR in three different dialects: MSA, Moroccan Arabic, and Egyptian Arabic. Since the language must be configured during inference, we conduct 3 separate inferences for each evaluation and report only the best result.


\subsection{Datasets}
The evaluation is conducted using a multi-dataset strategy to obtain robust scores for each model. We use only multi-dialectal datasets instead of separate datasets for each dialect. First, this is a more efficient method to cover a wider range of dialects and assess the model's generalizability.
Secondly, we consider it methodologically unsound to compare model performance on different dialects across datasets. For instance, if a model performs well on an MSA dataset but poorly on an Egyptian dataset, this does not necessarily indicate the model's poor generalization to the Egyptian dialect, as factors like recording conditions or other external variables may also impact the model's performance. The multi-dialectal datasets we have selected are as follows.

\textbf{SADA}: 
The Saudi Audio Dataset for Arabic (SADA) is a comprehensive speech dataset specifically designed for Arabic speech recognition. It contains over 600 hours of high-quality audio recordings, covering a wide range of dialects commonly used in Saudi Arabia as well as other commonly used dialects such as MSA and Egyptian. The dataset includes various speaking styles, accents, and diverse acoustic conditions, making it highly suitable for ASR training and evaluation. We employ its test set of 10.7 hours and 10 different dialects.

\textbf{Common Voice}: 
The Common Voice dataset is a large-scale, open-source collection of speech data to support research and development in ASR systems. The 18.0 version contains an Arabic subset with 91 hours of validated Arabic speech, including many common dialects such as Egyptian, Syrian, North African, etc. We use its test set of 12.6 hours and 25 different dialects. It should be noted that Common Voice shuffles the dataset splits from one release to another, the results may be biased for those models trained on previous versions due to the test data leakage, but according to the results, this will not significantly impact the overall benchmark.

\textbf{MASC}: 
The Massive Arabic Speech Corpus (MASC) is a comprehensive dataset that contains 1,000 hours of speech recordings from over 700 YouTube channels.  The dataset is multi-regional, multi-genre, and multi-dialect. Besides MSA, it covers more than 20 dialects that are commonly used in Syria, Egypt, Saudi Arabia, Jordan, etc. We evaluate the models with the test-clean and test-noisy subsets, the previous has 10.5 hours and 7 dialects, while the latter has 8.9 hours and 14 dialects.

\textbf{MGB-2}: 
The MGB-2 dataset is a large-scale speech corpus developed for the Arabic language, it contains approximately 1,200 hours of broadcast TV data, spanning a wide range of genres such as news, talk shows, and drama, recorded from 5 different Arabic dialects. In our benchmark, we adopted its 9.6-hour test set.

It should be noted that certain datasets were not included in our benchmark due to specific reasons. For example, the QASR dataset has been facing downloading issues.
An extension to the leaderboard will be made following the availability of an appropriate evaluation dataset.

\subsection{Metrics}
We compute and report both Word Error Rate (WER) and Character Error Rate (CER) in the leaderboard. For text normalization, we follow the work in \cite{textnorm}, removing all punctuation as well as the diacritics, and normalizing characters with Hamzas and Maddas. We also convert all Eastern Arabic numerals to Western Arabic numerals. Since certain models might support code-switching and the test sets contain very few code-switching samples, we kept the Latin characters and included code-switching errors in the overall error computation.

\section{Results and Analysis}

\subsection{Overall Leaderboard}
The overall leaderboard is shown in Table 1. For all the collected models, we performed zero-shot inference across five different multi-dialect test sets. We then averaged the WER and CER over the test sets and ranked the models based on the average WER. As can be seen from the results, Nvidia's Conformer-CTC-large-Arabic, combined with its released 4-gram language model, ranks first on the leaderboard with 25.71\% Average WER and 10.02\% Average CER. The Whisper large series and seamless-m4t model follow closely behind, while smaller Whisper models and fine-tuned self-supervised models rank last.

The results show that model rankings are closely aligned with the scale of the training datasets. For example, the top-ranked Conformer-CTC-large model was trained on 3,600 hours of labeled data, the middle-ranked whisper models were trained on more than 700 hours of labeled data, whereas the lower-ranked self-supervised models were only fine-tuned with less than 100 hours of labeled data, making them difficult to generalize well to all the test sets.



    




    



    

\begin{table}[htbp]
  \caption{Model performance (WER) under 3 different acoustic conditions using SADA dataset. The models are sorted based on the WER under the clean environment from lowest to highest.}
  \label{tab:environments}
  \centering
  \resizebox{\columnwidth}{!}{
    \begin{tabular}{c c c c}
    \toprule
    model & Clean$\downarrow$ & Noisy & Music \\
    \midrule
    nvidia/conformer-ctc-large-ar (lm) & \textbf{37,83\%} & \textbf{49,21\%} & \textbf{47,30\%} \\

    nvidia/conformer-ctc-large-ar (greedy) & 40,52\% & 51,64\% & 50,51\% \\

    openai/whisper-large-v2 & 51,21\% & 60,54\% & 61,76\%\\
    
    openai/whisper-large-v3 & 51,45\% & 58,47\%	& 58,71\%\\

    openai/whisper-large-v3-turbo & 52,96\% & 63,51\% & 66,08\% \\

    openai/whisper-large & 55,73\% & 70,26\% & 64,04\%\\

    facebook/seamless-m4t-v2-large & 56,98\% & 66,48\% & 64,69\% \\    

    openai/whisper-medium & 60,82\% & 73,23\% & 69,68\% \\
    
    asafaya/hubert-large-ft & 62,80\% & 70,72\% & 70,68\% \\

    facebook/mms-1b-all & 72,04\% & 80,92\% & 80,15\% \\

    whitefox123/w2v-bert-2.0-ft & 74,92\% & 79,88\% & 79,68\%\\

    openai/whisper-small & 77,48\% & 91,17\% & 95,47\% \\

    jonatasgrosman/wav2vec2-large-xlsr-ft & 82,62\% & 89,14\% & 89,35\% \\

    speechbrain/wav2vec2-large-ft & 85,21\% & 90,53\% & 90,34\% \\
    
    \bottomrule
    \end{tabular}
}
\end{table}

\subsection{Robustness}

Furthermore, we assessed the model's robustness under different acoustic conditions and across different dialects within the SADA dataset, the results are presented in Table 2 and Table 3. Under Noisy or Music environments, all models show a certain degree of performance degradation compared to the Clean environment. Similarly, Nvidia’s Conformer-CTC-Large model combined with its language model tops the list with the best WER under all environmental conditions, followed by the Whisper large series and self-supervised models.

As for the dialect variation, the results suggest that all models achieve their best results on MSA, but exhibit a significant decline when applied to dialects such as Egyptian and Khaliji. This effect is likely caused by the imbalanced distribution of dialects in most public datasets, with MSA generally representing a higher percentage than other dialects.

\begin{table}[htbp]
  \caption{ Model performance (WER) across 5 different dialects using SADA dataset. The models are sorted by the WER on MSA from lowest to highest.}
  \label{tab:environments}
  \centering
  \resizebox{\columnwidth}{!}{
  \begin{tabular}{c c c c c c}
    \toprule
    model & MSA$\downarrow$ & Egyptian & Hijazi & Khaliji & Najdi \\
    \midrule
    nvidia/conformer-ctc-large-ar (lm) & \textbf{19,23\%} & \textbf{40,97\%} & \textbf{36,96\%} & \textbf{48,23\%} & \textbf{36,34\%} \\

    nvidia/conformer-ctc-large-ar (greedy) & 21,56\% & 41,97\% & 39,84\% & 51,12\% & 38,70\% \\

    openai/whisper-large-v3 & 27,95\% & 59,28\% & 49,99\% & 59,92\% & 48,58\% \\

    openai/whisper-large-v3-turbo & 28,93\% & 63,01\% & 54,35\% & 65,04\% & 52,42\% \\

    openai/whisper-large-v2 & 29,13\% & 48,44\%	& 51,40\% & 63,06\% & 51,49\% \\

    openai/whisper-large & 31,89\% & 69,12\% & 59,67\% & 64,71\% & 58,01\% \\    

    openai/whisper-medium & 38,31\% & 55,79\% & 60,83\% & 71,30\% & 63,97\% \\

    asafaya/hubert-large-ft & 39,51\% & 74,47\% & 63,05\% & 73,12\% & 61,07\%\\

    facebook/seamless-m4t-v2-large & 39,87\% & 72,98\% & 54,02\% & 55,73\% & 54,89\%\\    

    facebook/mms-1b-all & 42,71\% & 81,94\% & 73,21\% & 83,43\% & 69,95\%\\

    openai/whisper-small & 49,41\% & 69,61\% & 86,62\% & 93,16\% & 87,69\% \\
    
    whitefox123/w2v-bert-2.0-ft & 51,13\% & 90,29\% & 73,60\% & 80,08\% & 75,62\% \\

    jonatasgrosman/wav2vec2-large-xlsr-ft & 61,51\% & 89,79\% & 84,48\% & 90,32\% & 82,16\% \\

    speechbrain/wav2vec2-large-ft & 66,06\% & 91,53\% & 86,50\% & 91,10\% & 85,26\% \\
    
    \bottomrule
  \end{tabular}
}
\end{table}

\begin{table}[htbp]
  \caption{Model performance (WER) across Speakers' age and gender, the models are sorted based on the WER for adult speakers from lowest to highest.}
  \label{tab:environments}
  \centering
  \resizebox{\columnwidth}{!}{
  \begin{tabular}{c c c c c}
    \toprule
    & \multicolumn{2}{c}{Age} & \multicolumn{2}{c}{Gender} \\
    model & Adult$\downarrow$ & Elderly & Female & Male \\
    \midrule
    nvidia/conformer-ctc-large-ar (lm) & \textbf{37,85\%} & \textbf{44,31\%} & \textbf{40,19\%} & \textbf{37,91\%} \\

    nvidia/conformer-ctc-large-ar (greedy) & 40,51\% & 44,41\% & 43,04\% & 40,46\% \\

    openai/whisper-large-v3 & 50,01\% & 55,28\% & 54,32\% & 49,60\% \\
    
    openai/whisper-large-v2 & 52,25\% & 52,51\% & 55,48\% & 52,03\% \\

    openai/whisper-large-v3-turbo & 53,74\% & 57,23\% & 55,79\% & 54,07\% \\

    facebook/seamless-m4t-v2-large & 54,46\% & 51,38\% & 48,84\%	& 55,53\% \\   
    
    openai/whisper-large & 58,06\% & 58,67\% & 62,40\% & 57,70\% \\
    
    asafaya/hubert-large-ft & 62,45\% & 70,36\% & 70,03\% & 61,64\% \\

    openai/whisper-medium & 62,47\% & 60,72\% & 68,62\% & 61,66\% \\
    
    facebook/mms-1b-all & 71,28\% & 80,92\% & 80,87\% & 70,12\% \\

    whitefox123/w2v-bert-2.0-ft & 73,99\% & 78,15\%	& 77,26\% & 73,70\% \\

    jonatasgrosman/wav2vec2-large-xlsr-ft & 82,66\% & 88,72\% & 88,68\% & 81,95\% \\

    speechbrain/wav2vec2-large-ft & 85,08\% & 89,95\% & 89,31\% & 84,63\% \\

    openai/whisper-small & 85,29\% & 92,10\% & 94,05\% & 84,59\% \\
    
    \bottomrule
  \end{tabular}
}
\end{table}

\subsection{Speaker Adaptation}

In order to ensure benchmark fairness and showcase evaluation bias, we also conduct an analysis of speaker adaptation on the SADA dataset, taking both the speaker's age and gender into account. The results indicate that for most of the models, elderly speakers generally have higher WER compared to adults, and female speakers tend to have slightly higher WER than male speakers, suggesting potential bias of age and gender in Arabic ASR model training.

\subsection{Efficiency and Resource Usage}

Given the importance of efficiency for production ASR models, we also evaluated the efficiency of each model on the SADA dataset. This includes the real-time factor (RTF) on GPU and the GPU memory usage during inference, covering both memory for model loading and average inference memory per sample. Instead of running tests on a full-load setup as in \cite{leaderboard}, we tried to avoid repeated work and used a minimal-load configuration. Namely, we used a batch size of 1 with an average audio duration of 6.25 seconds, which aligns more closely with real-world scenarios. The RTFs are measured on a single A100 40GB GPU with 5 warm-up runs.

The results reveal that, in conditions of small batch size and short audio duration, self-supervised models achieve higher efficiency, exhibiting lower RTF and reduced memory usage. In contrast, Whisper large models are significantly slower and consume more memory. The Conformer-CTC model, meanwhile, demonstrates impressive speed and requires minimal memory.

\begin{table}[htbp]
  \caption{Models' efficiency and resource usage on the SADA dataset with minimal batch load. The models are sorted by RTF from lowest to highest. \#Mem\_model refers to the memory usage (GB) with only the model loaded, while \#Mem\_model refers to the average inference-only memory (GB) per sample.}
  \label{tab:environments}
  \centering
  \resizebox{\columnwidth}{!}{
  \begin{tabular}{c c c c}
    \toprule
    model & RTF$\downarrow$ & \#Mem\_model & \#Mem\_infer \\
    \midrule
    jonatasgrosman/wav2vec2-large-xlsr-ft & \textbf{0,0064} & 1,21 & 0,14 \\

    facebook/mms-1b-all & 0,0137 & 3,69 & 0,14 \\

    speechbrain/wav2vec2-large-ft & 0,0137 & 1,23 & 0,14 \\

    asafaya/hubert-large-ft & 0,0142 & 1,23 & 0,14 \\

    nvidia/conformer-ctc-large-ar (greedy) & 0,0470 & 0,48 & \textbf{0,07} \\
    
    nvidia/conformer-ctc-large-ar (lm) & 0,0764 & 0,48 & \textbf{0,07} \\

    whitefox123/w2v-bert-2.0-ft & 0,1168 & 2,26 & 0,14 \\

    openai/whisper-small & 0,1791 & \textbf{0,48} & 0,12 \\

    openai/whisper-medium & 0,2314 & 1,43 & 0,15 \\

    facebook/seamless-m4t-v2-large & 0,2491 & 5,62 & 0,20 \\
    
    openai/whisper-large-v3-turbo & 0,3066 & 1,52 & 0,19 \\

    openai/whisper-large-v3 & 0,4335 & 2,88 & 0,24 \\


    openai/whisper-large & 0,4758 & 2,88 & 0,24 \\

    openai/whisper-large-v2 & 0,4789 & 2,88 & 0,24 \\

    \bottomrule
  \end{tabular}
}
\end{table}






    

    
    


    

\subsection{Discussions}
As a large-scale, comprehensive benchmark, this study has several limitations that are open to discussion. First, similar to \cite{leaderboard}, we tried to minimize the in-domain issue based on known information. However, given the limited language resources, we included the Common Voice dataset and some models fine-tuned on it. As our benchmark includes multiple datasets, this effect on the overall rankings is reduced. Additionally, the dataset inherently suffers from unavoidable attribute distribution imbalances, for example in the MGB-2, over 70\% of the samples represent the MSA dialect, which can potentially bias the results.

This benchmark is designed as a continuous and long-term project. We plan to regularly update the models and datasets alongside Arabic ASR progress to provide up-to-date references for the community. The leaderboard is hosted on HuggingFace\footnote {\url{https://huggingface.co/spaces/elmresearchcenter/open_universal_arabic_asr_leaderboard}} and the evaluation code is open-sourced on GitHub\footnote {\url{https://github.com/Natural-Language-Processing-Elm/open_universal_arabic_asr_leaderboard}}.

\section{Conclusions}
In this work, we implement the most comprehensive Arabic ASR benchmark to date, which includes an Open Universal Arabic ASR Leaderboard and a thorough analysis covering robustness, speaker adaptation, efficiency, and resource utilization. As a continuous effort, our work aims to provide the community with the latest references on model performance and develop a standard evaluation framework for Arabic ASR models.

\newpage








\bibliographystyle{IEEEtran}
\bibliography{mybib}

\end{document}